\documentclass[sigconf, nonacm=True]{aamas} 

\usepackage{balance} 

\usepackage{soul}
\usepackage{url}
\usepackage{graphicx}
\usepackage{amsmath}
\usepackage{booktabs}
\usepackage[ruled,longend]{algorithm2e}
\usepackage{textgreek}
\usepackage{tcolorbox}
\usepackage{wrapfig}
\usepackage{enumitem}
\usepackage{amsthm}
\usepackage{tikz}
\usetikzlibrary{positioning}

\setcopyright{ifaamas}
\acmConference[AAMAS '21]{Proc.\@ of the 20th International Conference on Autonomous Agents and Multiagent Systems (AAMAS 2021)}{May 3--7, 2021}{London, UK}{U.~Endriss, A.~Now\'{e}, F.~Dignum, A.~Lomuscio (eds.)}
\copyrightyear{2021}
\acmYear{2021}
\acmDOI{}
\acmPrice{}
\acmISBN{}

\acmSubmissionID{96}

\title[AAMAS-2021 Formatting Instructions]{SIBRE: Self Improvement Based REwards for Adaptive Feedback in Reinforcement Learning}

\author{Somjit Nath}
\affiliation{
  \institution{TCS Research}
  }

\author{Richa Verma}
\affiliation{
  \institution{TCS Research}
  }

\author{Abhik Ray}
\affiliation{
  \institution{BTS-Pilani (Goa)}}

\author{Harshad Khadilkar}
\affiliation{
  \institution{TCS Research}
  }

\begin{abstract}
We propose a generic reward shaping approach for improving the rate of convergence in reinforcement learning (RL), called \textbf{S}elf \textbf{I}mprovement \textbf{B}ased \textbf{RE}wards, or \textbf{SIBRE}. The approach is designed for use in conjunction with any existing RL algorithm, and consists of rewarding improvement over the agent's own past performance. We prove that SIBRE converges in expectation under the same conditions as the original RL algorithm. The reshaped rewards help discriminate between policies when the original rewards are weakly discriminated or sparse. Experiments on several well-known benchmark environments with different RL algorithms show that SIBRE converges to the optimal policy faster and more stably. We also perform sensitivity analysis with respect to hyper-parameters, in comparison with baseline RL algorithms.
\end{abstract}


\newcommand{\BibTeX}{\rm B\kern-.05em{\sc i\kern-.025em b}\kern-.08em\TeX}

\begin{document}


\pagestyle{fancy}
\fancyhead{}


\maketitle 


\section{Introduction} \label{sec:intro}

Reinforcement learning (RL) is useful for solving sequential decision-making problems in complex environments. Value-based \citep{mnih2015human,van2016deep}, actor-critic and its extensions \citep{DBLP:journals/corr/SchulmanLMJA15,DBLP:journals/corr/SchulmanWDRK17}, and Monte-Carlo methods \citep{guo2014deep} have been shown to match or exceed human performance in games. However, the training effort required for these algorithms tends to be high \citep{mnih2016asynchronous,silver2017mastering,pachockiopenai}, especially in environments with complex state-action spaces. One reason for slow learning is that complex environments typically have long episodes and a weak reward signal, requiring a large number of training episodes before initial successes are achieved.

In this paper, we propose a modification to the reward function (called \textbf{SIBRE}, short for Self Improvement Based REward) that aims to improve the rate of learning in episodic environments and thus addresses the problem of sample efficiency through reward shaping. SIBRE is a threshold-based reward for RL algorithms, which provides a positive reward when the agent improves on its past performance, and negative reward otherwise. We observe that this accelerates learning without requiring computationally expensive estimation of baselines \citep{greensmith2004variance}.

Furthermore, SIBRE can be used in conjunction with any standard RL algorithm: value or policy based, online or offline.

\textbf{Motivating applications in literature: }
Apart from game-based applications, reinforcement learning has been used in operations research problems \citep{zhang1995reinforcement}, robotics \citep{gu2017deep}, and networked systems \citep{o2010residential}. We expect SIBRE to be helpful in such scenarios, because it addresses challenges such as (i) lack of knowledge of optimal reward level, (ii) variation of optimal reward values across instances within the same domain, and (iii) weak differentiation between rewards from optimal and suboptimal actions. 

Similar approaches appear to have worked in literature on container loading \citep{verma2019reinforcement} and railway scheduling \citep{khadilkar2018scalable} problems, without being formally proposed or analysed. One study on bin packing does propose reward shaping explicitly, and is described below.

\textbf{Literature on formal reward shaping: }
The proposed approach (SIBRE) falls under the category of reward shaping approaches for RL, but with some key novelty points as described below. Prior literature has shown that the optimal policy learnt by RL remains invariant under reward shaping if the modification can be expressed as a potential function \citep{ng1999policy}. Other studies such as \cite{badnava2019new} have used potential functions for transfer learning. While the concept is valuable, designing a potential function for each problem could be a difficult task. Another potential-based reward approach \cite{10.5555/2886521.2886690} adds a heuristic-based reward as potential function which needs to be specified by the user as per the environment, making it not directly usable with any RL setting. In \cite{NIPS2019_9225}, the authors use a distance-based reward for improving diverse exploration. We cover this method in detail in Section \ref{sec:sr}. In \cite{DBLP:journals/corr/abs-1806-07857}, the authors suggest a reward shaping approach which works for delayed reward tasks however, training it is computationally expensive as compared to SIBRE because of an additional LSTM network. Also, SIBRE works with both sparse and dense reward environments. 

A study by \cite{laterre2018ranked}, shows improved performance on a bin packing task under a ranked reward scheme, where the agent's reward is based on its recent performance. The reward signal is binary ($\pm1$), and is based on a comparison with the 75th percentile of recently observed rewards. These binary rewards are used as targets for value estimation. Similar intuition is also used in Self-Imitation Learning (SIL) \cite{oh2018selfimitation} even though it is not a reward shaping algorithm. Here, the authors develop an off-policy Actor Critic algorithm which learns from the past and tries to reproduce the good actions which helps in solving hard exploration tasks. While SIBRE is conceptually similar, the key differences from these as well as other reward shaping approaches are,
\begin{itemize}
    \item Our reward modification is computationally light (simple average).
    \item It can be used to improve the sample efficiency of any RL algorithm.
    \item We prove that SIBRE converges in expectation to the same policy as the original algorithm.
    \item We empirically observe faster convergence with lower variance on a variety of benchmark environments, with multiple RL algorithms, and in addition to other reward shaping approaches.
\end{itemize}
\textbf{Note:} SIBRE is conceptually similar to REINFORCE with a baseline \cite{reinforce}, which is a policy gradient method. Unlike REINFORCE which has a constant baseline, SIBRE uses an auto-adaptive baseline, and since it is a reward shaping approach, SIBRE can be used in \textit{any} RL setting.



We define SIBRE formally in Section \ref{sec:method}, derive the convergence characteristics in Section \ref{sec:convergence}, and describe experimental results in Section \ref{sec:results}.


\section{Description of Methodology} \label{sec:method}

Consider an episodic Markov Decision Process (MDP) specified by the tuple $<\mathcal{S}, \mathcal{A}, \mathcal{R}, {P}>$ \citep{Sutton:2018}, where $\mathcal{S}$ is the state space, $\mathcal{A}$ is the action space, $\mathcal{R}$ is the set of possible rewards, and $P$ is the transition function. We assume the existence of a reinforcement learning algorithm for learning the optimal mapping $\mathcal{S}\rightarrow\mathcal{A}$. We use Q-Learning \citep{Watkins1992} for ease of explanation in this section, but similar arguments can be developed for policy and actor-critic based algorithms. Typically, the reward structure is a natural consequence of the problem from which the MDP was derived. For example, in the popular environment Gridworld \citep{gym_minigrid}, the task is to navigate through a 2-D grid towards a goal state. The most common reward structure in this problem is to provide a small negative step reward for every action that does not end in the goal state, and a large positive terminal reward for reaching the goal state. It follows that the value of the optimal reward depends on both the values of the step and terminal rewards, as well as on the size of the grid. In this paper, we retain the original step rewards $R_k$ for time step $k$ within the episode, but replace the terminal reward for episode $t$ by a baseline-differenced value of the total return $G_t : \mathcal{S}, \mathcal{A}, \mathcal{S} \,\to\, {\rm I\!R}$:
\begin{equation}
    r_{k,t}(s_{k},a_{k},s_{k+1}) = \begin{cases}
        G_{t} - \rho_{t}, & s_{k+1} \in \mathcal{T} \\
        R_{k}, & \text{otherwise}
    \end{cases}
    \label{eq:sibre}
\end{equation}
where $k$ is the step within an episode, $t$ is the number of the episode, $\mathcal{T}$ is the set of terminal states, $G_t$ is the return for episode $t$, and $\rho_t$ is the performance threshold at episode $t$. Note that the return $G_t$ is based on the original reward structure of the MDP. If the original step reward at $k$ is $R_k$, then $G_t = \sum R_{k}$. The net effect of SIBRE is to provide a positive terminal reward, if $G_t \geq \rho_t$ and negative otherwise, which gives the notion of self-improvement. For the purposes of the subsequent proof, we assume that a number $x$ of episodes is run after every threshold update, allowing the q-values to converge with respect to the latest threshold value. Note that $x$ can be a different number from one update to another. Once the q-values have converged, the threshold can be updated using the relation, 
\begin{equation*}
\rho_{t+1} = \begin{cases} \rho_{t} + \beta_{t}(\sum_{y=t-x+1}^{t}\frac{G_{y}}{x} - \rho_{t}) & \text{if updating q-values} \\ \rho_t & \text{otherwise} \end{cases},
\end{equation*}
where $\beta_{t} \in (0,1)$ is the step size and is assumed externally defined according to a fixed schedule.

Though we defined this for episodic problems, SIBRE can also be extended to continuous problems as explained in Section \ref{sec:continuing}).

\textbf{Training process:} The q-values are trained after every episode $t$, while the threshold is updated after $x$ episodes once the q-values have converged. If the initial threshold value is very low, it is fairly easy for the algorithm to achieve the positive terminal reward, and a large proportion of state-action pairs converge to positive q-values. During the next threshold update, the high average returns since the last update result in an increase in the value of $\rho_t$. The threshold thus acts as a lagging performance measure of the algorithm over the training history. 

The algorithm is said to have converged when both the threshold and the q-values converge. The original rewards $R$ only affect the returns $G$, which in turn are used to update the threshold $\rho$ for SIBRE. At the end of each episode, the current return and threshold values are used to compute the new rewards $r$, which implicitly or explicitly drive the policy $\pi$. The pseudo-code for the algorithm with Q-learning is shown in Algorithm \ref{alg:SIBRE}. The algorthm can be similarly integrated with any other RL algorithm.
%
            

\begin{algorithm}
Algorithm parameters: step size $\alpha \in (0, 1]$, small $\epsilon > 0$, $\beta \in (0,1)$\; Threshold Update after K episodes assuming Q-values converge after K episodes\;
Initialize $Q(s, a)$, for all $s \in \mathcal{S}^+, a \in \mathcal{A}(s)$, arbitrarily except that $Q(\mathrm{terminal}, \cdot) = 0$\; $\rho = 0$\;
\ForEach{episode}{
    Initialize S\;
    $G$ = 0\;
    $episodes$ = 1\;
    \ForEach{step of episode}{
        Choose $A$ from $S$ using policy derived from $Q$ (e.g., \textepsilon-greedy)\;
        Take action $A$, observe $R$, $S'$\;
        $G = G + R$\; 
        \If{S $\in \mathrm{terminal}$}
        {$R = G - \rho$\;
        $episodes = episodes + 1$\;
        \If{$episodes\bmod K = 0$}{
        $\rho = \rho + \beta(G-\rho)$\;}
        }
        $Q(S, A) \leftarrow Q(S, A) + \alpha [R + \gamma \max_a Q(S', a) - Q(S, A)]$\;
        $S \leftarrow S'$\;
    }
}
\caption{Illustration of SIBRE using Q-learning as example}
\label{alg:SIBRE}
\end{algorithm}

\section{Proof of Convergence}\label{sec:convergence}
After every update to the threshold $\rho$, we assume that the chosen reinforcement learning algorithm is allowed to converge using the modified SIBRE rewards $r$. Further, we assume that the chosen algorithm in its default form is known to converge to the optimal policy. For example, in Q-Learning it is known that $Q$ values converge to $Q^{*}$ under mild conditions \citep{jaakkola}. Let $\rho^{*} = \mathbb{E}_{\pi^{*}}[G] = V^{*}(s_{0})$ be the maximum expected return from the start state for the original reward structure. The following theorem shows that the algorithm with SIBRE-defined rewards also converges in expectation to the same return. Note that we do not make any assumptions about the form of the RL algorithm: on- or off-policy, value- or policy-based.

\textbf{Theorem 1}: An RL algorithm with known convergence properties, still converges to the optimal policy when used in conjunction with SIBRE. 



\textbf{Proof:} We prove the conjecture by considering the following three cases for the current threshold $\rho_t$ with respect to the optimal threshold $\rho^*$: $\mathbb{E}[\rho_{t}] < \rho^{*}$, $\mathbb{E}[\rho_{t}] > \rho^{*}$ and $\mathbb{E}[\rho_{t}] = \rho^{*}$. The crux of the proof is to show that the expectation of $\rho_t$ moves towards $\rho^*$ (or remains at $\rho^*$) in all the three cases. Also, since $\rho^{*}$ is defined as the maximum expected return from the start state for the original reward structure,  the optimal policy for the new reward structure, SIBRE, when $\rho = \rho^{*}$ is the optimal policy for the original MDP. The full proof is given in supplementary material, along with some observations. 

\textit{Case 1}: $\mathbb{E}[\rho_{t}] < \rho^{*}$ \\
There exists a policy for which $\mathbb{E}[G] = \rho^{*}$, therefore, there exist policy for which $\mathbb{E}[G_{t}] > \mathbb{E}[\rho_{t}]$.
If we let the RL algorithm converge, $\mathbb{E}[G_{t}] > \mathbb{E}[\rho_{t}]$
\vspace{-5mm}
\begin{proof}[\unskip\nopunct]\renewcommand{\qedsymbol}{}
    \begin{align}
        \nonumber \rho_{t+1} &= \rho_{t} + \beta_{t}(G_{t} - \rho_{t})\\
        \nonumber \mathbb{E}[\rho_{t+1}] &= \mathbb{E}[\rho_{t}]+ \beta_{t}\mathbb{E}[G_{t} - \rho_{t}]\\
        \mathbb{E}[\rho_{t+1}] &> \mathbb{E}[\rho_{t}]
    \end{align}
    \begin{center}
        \text{\{Since $\beta_{t}>0$ and $\mathbb{E}[G_{t}] > \mathbb{E}[\rho_{t}]$\}}
    \end{center}
    \vspace{-3mm}
    \begin{align}
        \nonumber \rho_{t+1} &= (1-\beta_{t})\rho_{t} + \beta_{t}G_{t} \\
        \nonumber \mathbb{E}[\rho_{t+1}] &= (1-\beta_{t})\mathbb{E}[\rho_{t}] + \beta_{t}\mathbb{E}[G_{t}] \\
        \nonumber &< (1-\beta_{t})\rho^{*} + \beta_{t}\rho^{*}\\
        \mathbb{E}[\rho_{t+1}] &< \rho^{*}
    \end{align}
    \begin{center}
        \text{\{Since $\rho^{*} \geq \mathbb{E}[G_{t}]$ by definition\}}
    \end{center}
\end{proof}
\vspace{-5mm}
From (1) and (2):
\begin{equation}
    \mathbb{E}[\rho_{t}] < \mathbb{E}[\rho_{t+1}] < \rho^{*}
\end{equation}

\textit{Case 2}: $\mathbb{E}[\rho_{t}] > \rho^{*}$ \\[5pt]
There exists no policy for which $\mathbb{E}[G_{t}] \geq \mathbb{E}[\rho_{t}]$ since by definition of $\rho^{*}$, it is the maximum expected return.
If we let the RL algorithm converge, $\mathbb{E}[G_{t}] < \mathbb{E}[\rho_{t}]$
\vspace{-5mm}
\begin{proof}[\unskip\nopunct]\renewcommand{\qedsymbol}{}
    \begin{align}
    \nonumber\rho_{t+1} &= \rho_{t} + \beta_{t}(G_{t} - \rho_{t})\\
    \nonumber\mathbb{E}[\rho_{t+1}] &= \mathbb{E}[\rho_{t}] + \mathbb{E}[\beta_{t}(G_{t} - \rho_{t})]\\
    \mathbb{E}[\rho_{t+1}] &< \mathbb{E}[\rho_{t}]
    \end{align}
    \begin{center}
        \text{ \{Since $\beta_{t}>0$ and $\mathbb{E}[G_{t}] < \mathbb{E}[\rho_{t}]$\} }
    \end{center}
\end{proof}
\vspace{-5mm}
\textit{Case 3}: $\mathbb{E}[\rho_{t}] = \rho^{*}$ \\[5pt]
There exists a policy for which $\mathbb{E}[G] = \rho^{*}$, therefore, for the same policy $\mathbb{E}[G_{t}] = \mathbb{E}[\rho_{t}] = \rho^{*}$.
If we let the RL algorithm converge, $\mathbb{E}[G_{t}] = \mathbb{E}[\rho_{t}]$

\begin{proof}[\unskip\nopunct]\renewcommand{\qedsymbol}{}
    \begin{align}
    \nonumber\rho_{t+1} &= \rho_{t} + \beta_{t}(G_{t} - \rho_{t})\\
    \nonumber\mathbb{E}[\rho_{t+1}] &= \mathbb{E}[\rho_{t}] + \mathbb{E}[\beta_{t}(G_{t} - \rho_{t})] \\
    \mathbb{E}[\rho_{t+1}] &= \mathbb{E}[\rho_{t}] = \rho^{*}
    \end{align}
    \begin{center}
        \text{ \{Since $\mathbb{E}[G_{t}-\rho_{t}]=0$\} }
    \end{center}
\end{proof}
\vspace{-5mm}
Hence proved,
\begin{align*}
    \mathbb{E}[\rho_{t}] < \rho^{*} & \implies  \mathbb{E}[\rho_{t}] < \mathbb{E}[\rho_{t+1}] < \rho^{*} \\
    \mathbb{E}[\rho_{t}] > \rho^{*} & \implies \mathbb{E}[\rho_{t+1}] < \mathbb{E}[\rho_{t}] \\
    \mathbb{E}[\rho_{t}] = \rho^{*} & \implies \mathbb{E}[\rho_{t+1}] = \mathbb{E}[\rho_{t}] = \rho^{*}
\end{align*}

Therefore,  $\rho \to \rho^{*}$ and the optimal policy for the new reward structure, SIBRE, when $\rho = \rho^{*}$ is the optimal policy for the original MDP since by definition, the optimal policy is one that attains the maximum expected reward.

\subsection{Notes on the Proof of Theorem 1}
\begin{itemize}
    \item A step in the direction of optimality for the new reward structure $r_{k,t}$, given by SIBRE, is also a step in the direction of optimality for the original reward structure $\mathcal{R}$.
    \item In order to observe the convergence characteristics as described above, we do not necessarily need to let the RL algorithm converge to the final policy after each threshold update. We only need sufficient training to ensure,
    \begin{center}
        $\mathbb{E}[G_{t}] > \mathbb{E}[\rho_{t}]$ for $\mathbb{E}[\rho_{t}] < \rho^{*}$\\
        $\mathbb{E}[G_{t}] < \mathbb{E}[\rho_{t}]$ for $\mathbb{E}[\rho_{t}] > \rho^{*}$\\
        $\mathbb{E}[G_{t}] = \mathbb{E}[\rho_{t}]$ for $\mathbb{E}[\rho_{t}] = \rho^{*}$
    \end{center}
    \item In practical use, we found that training once after every episode also shows similar convergence characteristics. We use this approximation for all the results reported later in this study.
\end{itemize}

\textbf{Intuition: }
Our hypothesis is that the rewards under SIBRE, as defined in (\ref{eq:sibre}), help RL algorithms discriminate between \textit{good} and \textit{bad} actions more easily. In addition to absolute performance (measurable within an episode), SIBRE provides a relative performance feedback (over multiple episodes), thus enriching the reward signal. The effect is similar to that of baselines in policy gradient algorithms \citep{greensmith2004variance}, but can be generically applied to all reinforcement learning algorithms, including value-based methods. The effect is particularly noticeable in environments with a weak reward signal, or in environments where the measurable difference in outcomes (rewards) is small. In both cases, the terminal reward in (\ref{eq:sibre}) becomes significant in proportion to the total rewards over the episode. Section \ref{sec:results} provides empirical support for this reasoning.

\textbf{Practicalities: }
Observe that a step towards improvement for the SIBRE-defined reward $r$ is also a step towards improvement for the original reward. Therefore, it turns out that SIBRE works well (improves reward consistently) even if we don't wait for full convergence at every threshold. In the rest of this paper, we update the threshold after each episode.

\section{Experiments and Results} \label{sec:results}


To validate our approach and evaluate its effectiveness, we test SIBRE along with a variety of other value based RL algorithms like Q-learning \citep{Sutton:2018} and Rainbow \citep{DBLP:journals/corr/abs-1710-02298}, and policy-based (offline and online) algorithms like A2C \citep{mnih2016asynchronous} and PPO \citep{schulman2017proximal}. 

\subsection{Environments}
In this section, we describe the environments and base RL algorithms used to test SIBRE along with key hyper-parameters. A complete set of hyper-parameter settings are given in the supplementary material. The sensitivity of SIBRE to $\beta$ and the reasoning behind choosing each $\beta$ is explained in Section \ref{sec:beta}.

\begin{itemize}[leftmargin=*]
    \item \textbf{Door \& Key environment}: We use a variable sized Gridworld environment \citep{gym_minigrid} with a negative step reward and a positive terminal reward for reaching the goal. The environments are partially observable with the agent seeing only a portion of the environment. This environment has a key that the agent must pick up to unlock a door and then get to the gold. This environment is challenging to solve because there is no additional reward for picking up the key. 
    The initial position of the agent, the position of gold and that of the key are all randomly set. An episode terminates after the agent reaches the gold or after 1000 time-steps are elapsed, whichever happens earlier. 
    
    There is a $-0.1$ penalty for every step and a $+4.0$ reward for reaching the gold. There is no intermediate reward for picking the key or opening the door which makes the task challenging.
    
    In this environment, we used A2C with and without SIBRE on 6x6 Door \& Key environment. Both the algorithms were trained for 1.8 Million frames over 10 runs. We started of with an initial value of $\beta = 0.001$ and increased it linearly to 0.1 after every 10\% of the total number of episodes.
    
    
    \item \textbf{Multi-room environment}: This gridworld environment has a series of connected rooms with doors that must be opened to get to the next room. The final room has the gold that the agent must get to. This environment is also challenging particularly with increasing number of rooms.
    The initial position of the agent and that of the gold are set randomly. An episode terminates after the agent reaches the gold or after 1000 time-steps are elapsed, whichever happens earlier. 
    Initially, there is a $-0.1$ penalty for every step and a $+4.0$ reward for reaching the gold. There is no intermediate reward for moving from one room to the other which increases the complexity of the task.
    A2C and A2C with SIBRE were tested on Two Rooms, for 1 Million frames over 10 runs, with SIBRE using the same $\beta$ schedule as DoorKey.

    \item \textbf{FrozenLake}: FrozenLake is a stochastic world setting, where the agent needs to traverse a grid (4x4). Some of the states have slippery ice which are safe to walk on, some have holes where the episode terminates with no reward. One of the states has a goal which provides a terminal reward of +1. Additionally, the stochasticity is induced in the environment because of the slippery ice, where the state the agent ends up in partially depends on the chosen action.
    For this environment, we trained an agent with tabular-Q learning. The agent was trained for 10000 episodes with a turn-limit of 100 steps per episode in which case the agent terminates with a reward of 0. We used $\epsilon$-decay style exploration and the same $\beta$-schedule.
    
    \item \textbf{CartPole}: CartPole is an OpenAI gym environment \cite{brockman2016openai} where the goal of the agent is to balance a pole on a stick. The only actions of the environment are to move the stick either left or right. The pole falls over if the angle between the normal to the stick and the pole exceeds $15^{\circ}$.
    
    We used a non-episodic version of this environment, as explained in Section \ref{sec:continuing}. We trained DQN \cite{mnih2015human} and DQN with SIBRE (scheduled $\beta$) for 1 Million Frames across 10 independent runs. For SIBRE, we used a $\beta$-schedule as the previous environments, except $\beta$ was updated at every 10\% total number of steps and not episodes, as this is a continuing task.
    
    \item \textbf{MountainCar}: MountainCar is a continuous action-space environment where the goal of the agent is to climb up a mountain where it gets a positive reward of 100. The goal is also to get to the top with the least force utilised which is subtracted from terminal reward. 
    
    For this environment, we trained A2C and A2C with SIBRE ($\beta = 0.1$) for 50 episodes across 50 independent runs.

    \item \textbf{Pong}: In Pong, the agent has to defeat the opponent by hitting a ball with a paddle and making the opponent miss the ball. The agent gets a reward of +1 every-time the opponent misses and a reward of -1 every-time the agent misses the ball. The episode terminates after the agent gets to $\pm{21}$. 
    
    Since we used policy-based methods (A2C) for our previous experiments, for Atari games, we experimented using a value-based method, Rainbow \citep{DBLP:journals/corr/abs-1710-02298}, which shows good performance across almost all Atari games.
    
    For Pong, we used Rainbow without multi-step updates (n=1). The agent was trained for 1 Million frames and the results across the 5 runs are plotted in Fig. \ref{fig:gym} (b). Also, the remaining hyper-parameters used are same as in \citep{DBLP:journals/corr/abs-1710-02298} with SIBRE using $\beta = 0.1$. 
    
    \item \textbf{Freeway}: In Freeway, the agent has to cross a road with traffic without hitting the car and every-time it crosses the road the agent gets a reward of +1 and it is moved back if it hits a car. This is also a relatively sparse reward setting.
    
    To solve this environment, we use Rainbow with same original settings as \cite{DBLP:journals/corr/abs-1710-02298} and we run it for 30 Million Frames. We plot the mean episode rewards every 125k frames. For Freeway to work well without SIBRE, we clipped the rewards to (-1,+1) whereas with SIBRE we passed the original rewards to SIBRE. All the remaining hyper-parameters are same as \cite{DBLP:journals/corr/abs-1710-02298}, and we used $\beta = 0.1$ for SIBRE. 
    
    \item \textbf{Venture}: Venture is a maze-navigation game, which is difficult to solve because of the sparsity of rewards. The agent has to navigate through a maze with enemies and escape within a certain time to avoid being killed. The reward is dependent on the time it takes to exit the maze. 
    
    Rainbow \citep{DBLP:journals/corr/abs-1710-02298} does pretty well in this game as opposed to many Deep Reinforcement Learning algorithms which struggle to solve the task. Both the algorithms were run for 120 Million frames. In this game too, we used the same hyper-parameters as used for Freeway. Rainbow used reward clipping to improve performance. SIBRE however, did not use any reward clipping and we ran it with $\beta = 0.1$.

    \item \textbf{2D Point Maze Environment} 
    In this environment \citep{NIPS2019_9225}, the agent has to navigate a 2D maze and get to a goal. The initial position of the agent and that of the goal are sampled randomly from a pre-defined area. An episode terminates after the agent reaches the goal or after 50 time-steps are elapsed, whichever happens earlier.
    
    We combine SIBRE with the reward shaping technique of \citep{NIPS2019_9225} and train it on a fully continuous 2D point maze environment used by the authors with $\beta = 0.002$ obtained after hyper-parameter search. We also run it with the same $\beta$-schedule where we increase $\beta$ linearly after every 10\% of the number of episodes. We use only terminal rewards. We do not change the hyper-parameter values in the code provided by the authors to combine it with SIBRE. Both the algorithms were trained for 20K episodes over 3 runs. 
    
    SIBRE further reshapes the distance-to-goal reward mentioned in their work, which uses PPO \citep{schulman2017proximal} as the underlying RL algorithm.

\end{itemize}

\subsection{Accelerating Learning}
In this section, we portray the most significant results of the paper, where SIBRE combined with base RL algorithms can speed up convergence and thus improve performance. 

\begin{figure*}[h]
\centering
\includegraphics[width=0.48\textwidth]{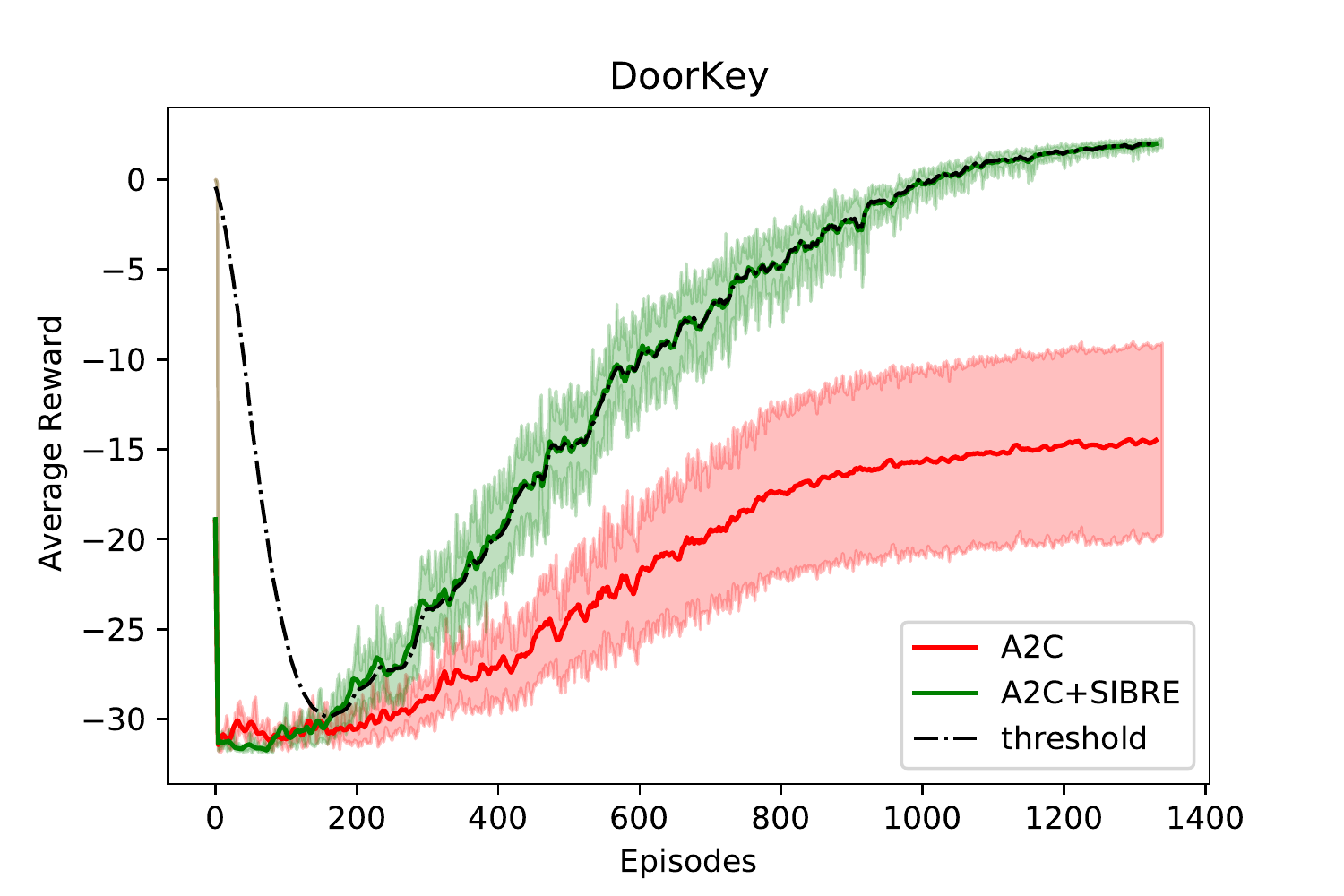} 
\includegraphics[width=0.48\textwidth]{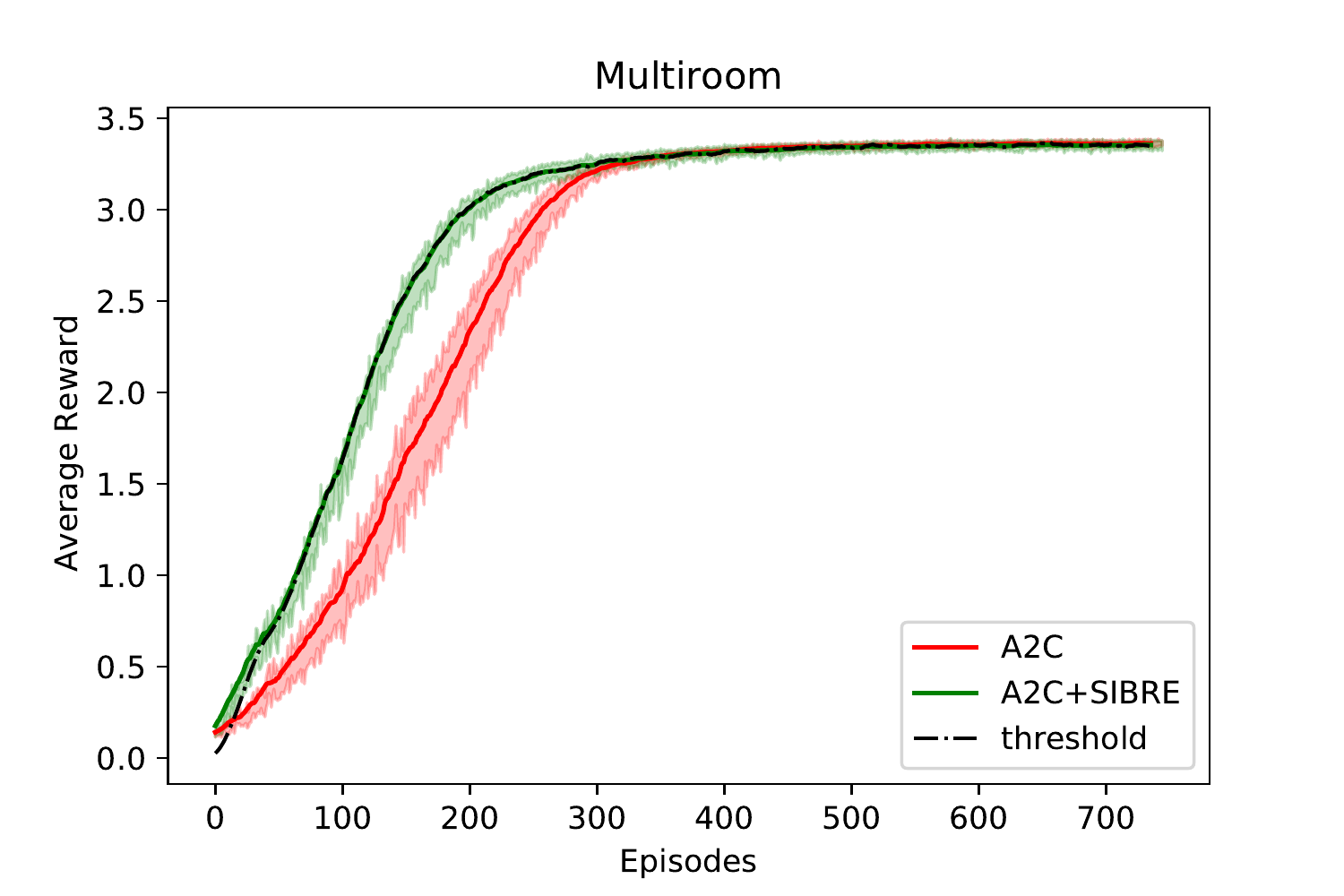} 
\caption{(a) SIBRE+A2C vs A2C on 6x6 Door \& Key Environment (b) Two Room Environment. }
\label{fig:grid}
\end{figure*}

\textbf{Gridworld}: Figure \ref{fig:grid} illustrates the mean of the rewards across 10 runs along with standard errors. In both the gridworld environments, we observe accelerated learning with SIBRE, which makes A2C+SIBRE much more sample efficient than normal A2C as shown in Fig. \ref{fig:grid}. The variance across runs, particularly in Fig. \ref{fig:grid} (a) is lesser, showing it consistently performs better for all the runs. The threshold value plots (dotted black lines) are basically a lagging value of the rewards, and stay at the same value upon convergence as seen in Fig. \ref{fig:grid} (b). 

For both these environments, we used a $\beta$-scheduling approach as explained in Section \ref{sec:beta}. However, from our experience, we found $\beta = 0.1$ to be a good starting point which performs well more or less for all environments.

\begin{figure*}[h]
\centering
\includegraphics[width=0.48\textwidth]{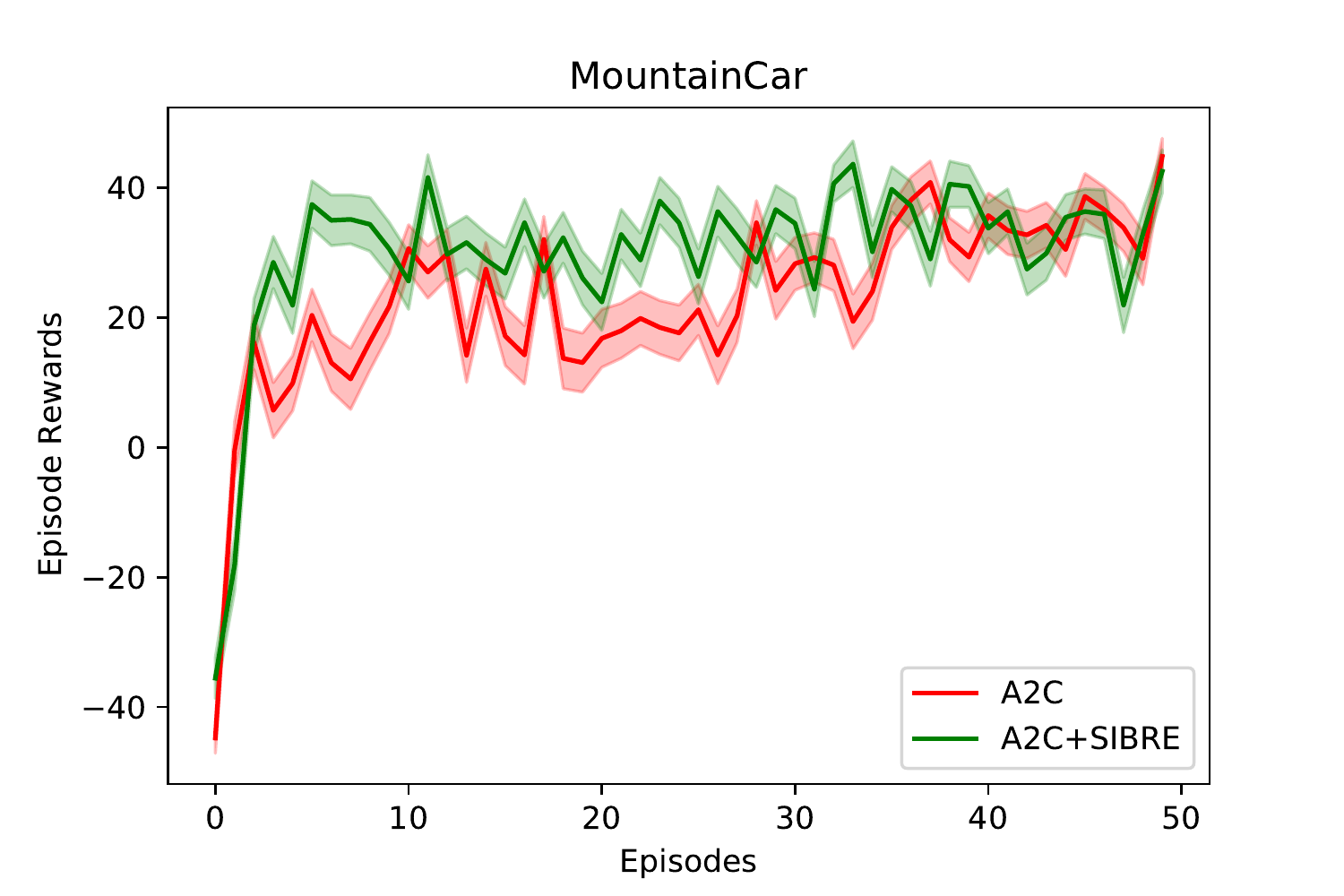} 
\includegraphics[width=0.48\textwidth]{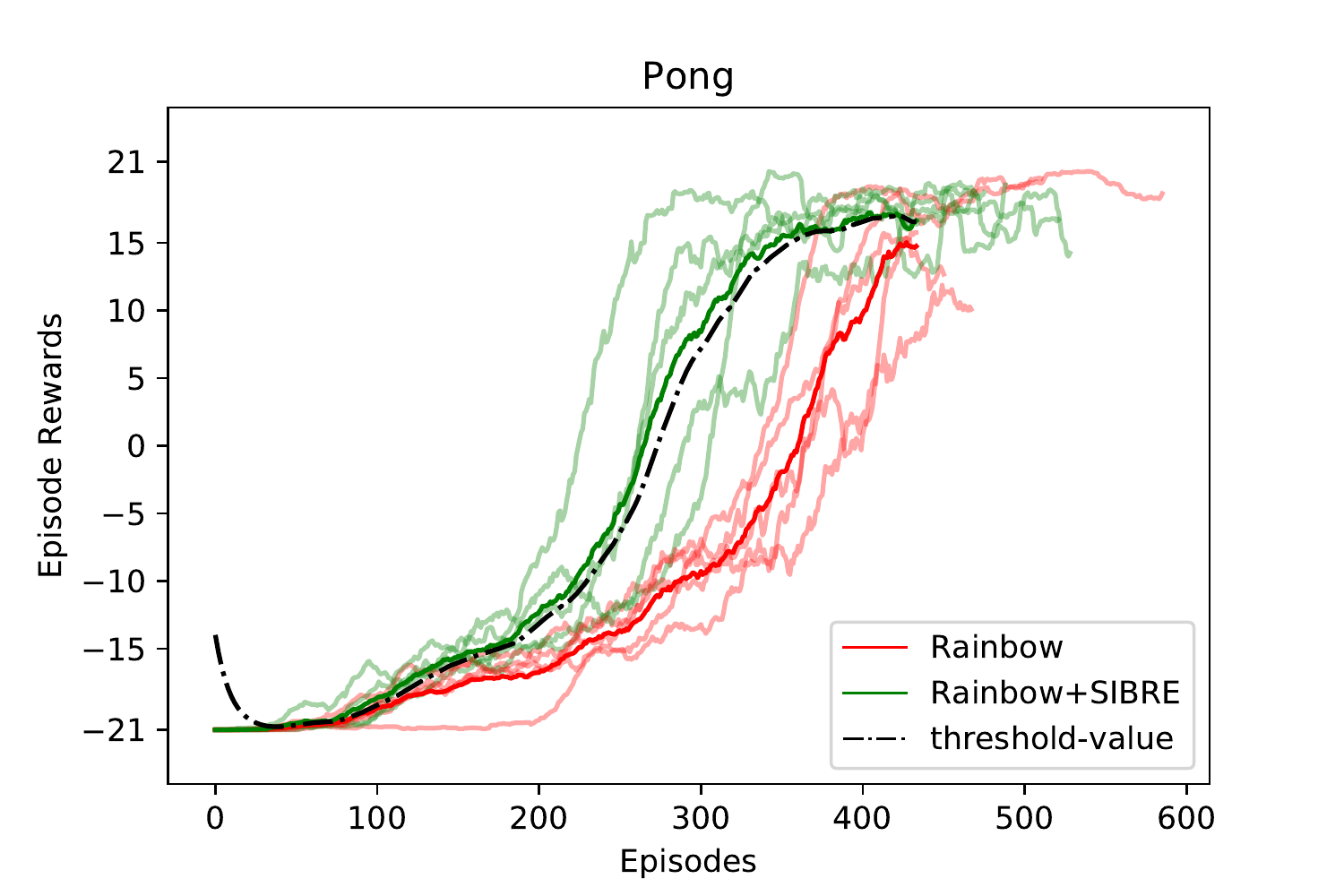} 
\caption{(a) SIBRE+Q-learning vs Q-learning on MountainCar (b) SIBRE+Rainbow vs Rainbow on Pong}
\label{fig:gym}
\end{figure*}

\textbf{OpenAI Gym Environments}: For both the environments, we did not find such major performance enhancements, mainly due to the fact that these environments are easier to solve. With tabular Q-learning on FrozenLake, we found both to perform almost similarly (plot in supplementary material). For MountainCar however, we observe an accelerated learning, but both algorithms learn sufficiently well, shown in Fig. \ref{fig:gym} (a).

\textbf{Atari}: Across all the 5 runs in pong in Fig. \ref{fig:gym} (b), Rainbow with SIBRE reaches optimal policy much faster than Rainbow and even the worst performing run with SIBRE is doing better than the best performing run without SIBRE, which signifies the improvement of using a threshold based reward structure to accelerate performance. However, asymptotically both reach optimal performance.

\begin{figure*}[h]
\centering
\includegraphics[width=0.48\textwidth]{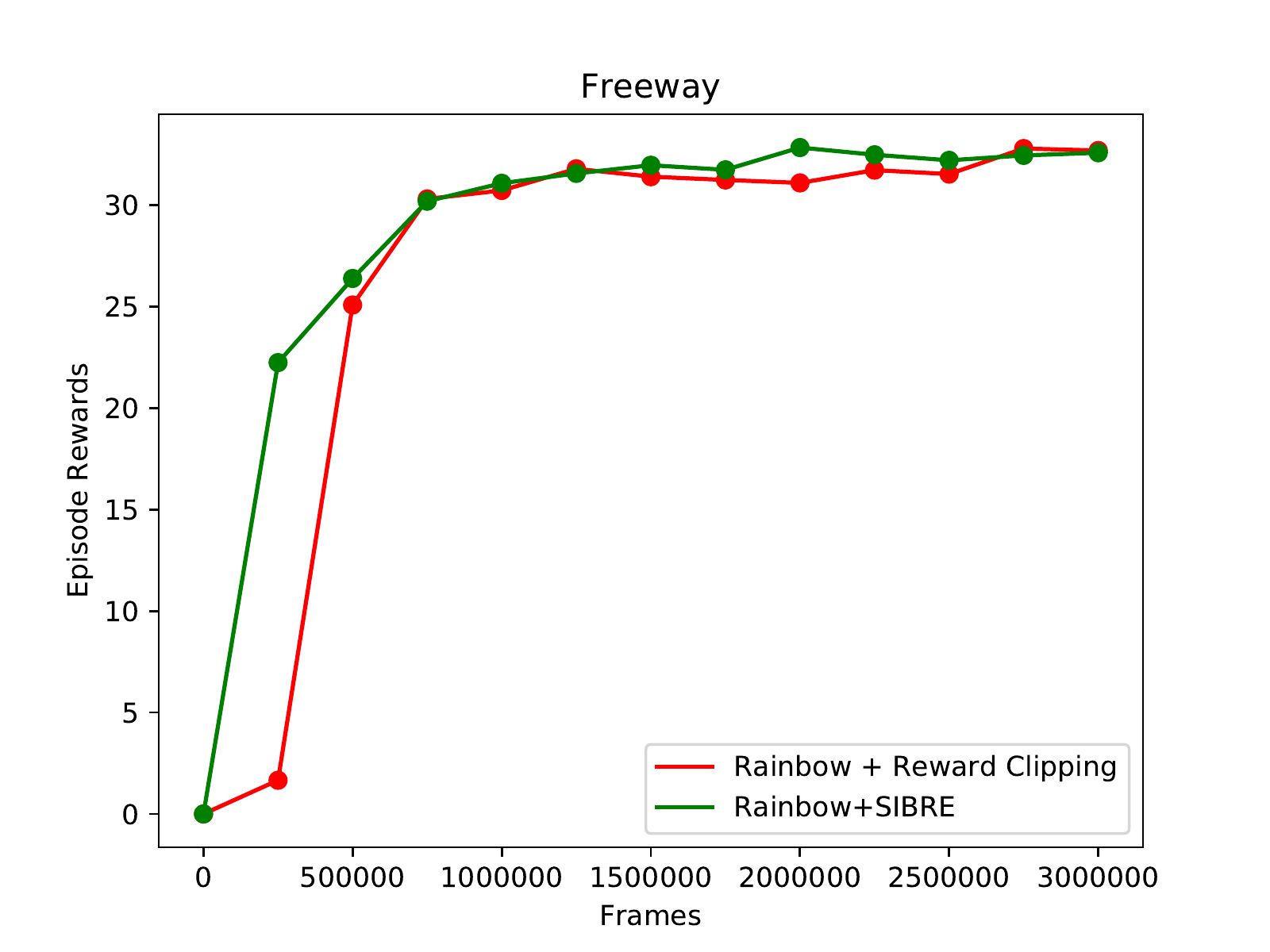} 
\includegraphics[width=0.48\textwidth]{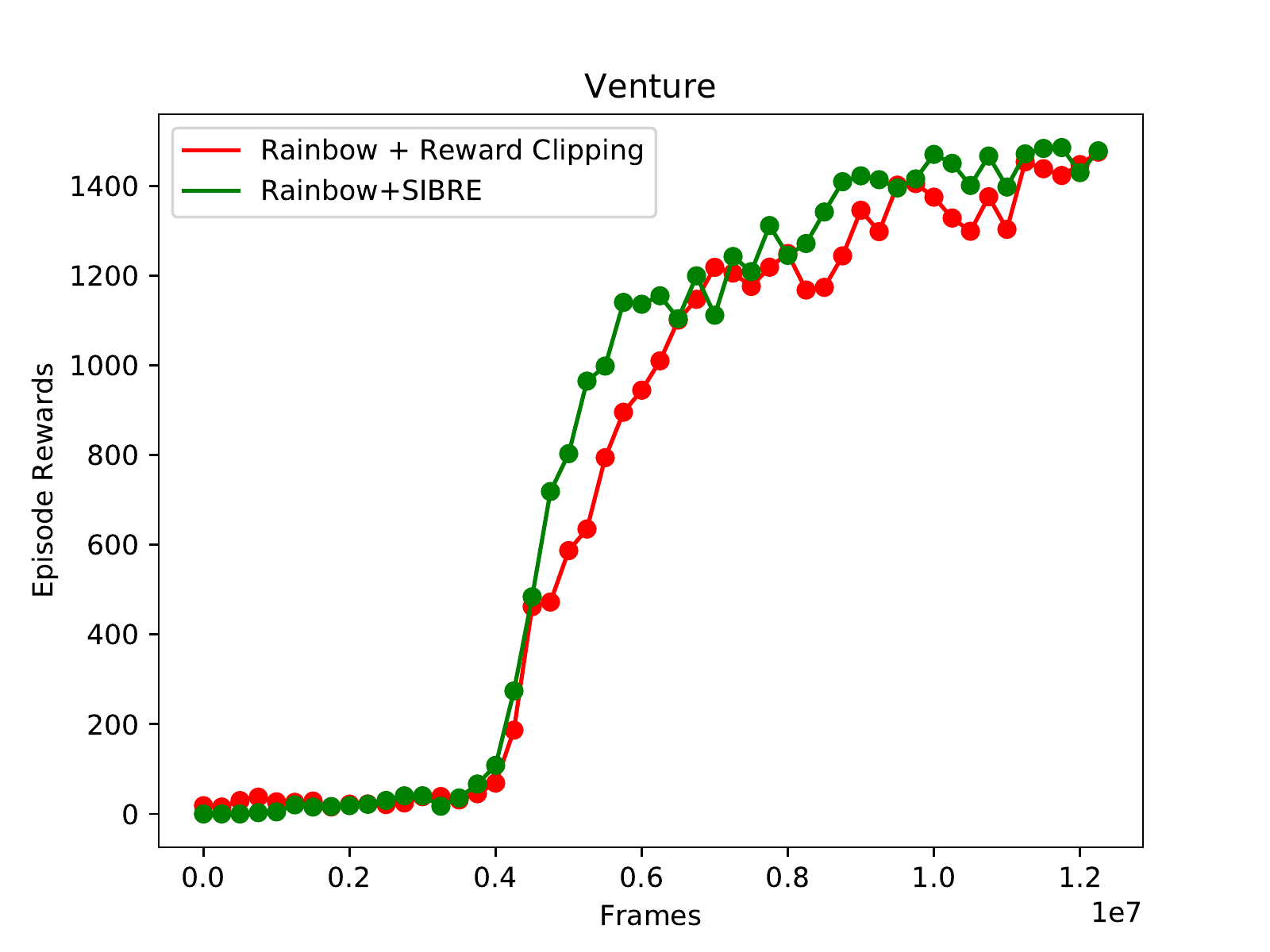} 
\caption{(a) SIBRE+Rainbow vs Rainbow  on Freeway  (b) SIBRE+Rainbow vs Rainbow on Venture}
\label{fig:atari}
\end{figure*}

For Freeway and Venture, Fig. \ref{fig:atari}, we see a similar story, with addition of SIBRE to Rainbow making it more sample efficient.

\subsection{$\beta$-Sensitivity}
\label{sec:beta}
SIBRE introduces an additional parameter $\beta$ whose value changes the step-size for updating the threshold values. If we update the value after every episode, then the update equation is:
$$\rho_{t+1} = \rho_t + \beta(G_t-\rho_t)$$ where $G_t$ is the return of the episode. This can be rewritten as:
$$\rho_{t+1} = \rho_t + \beta(G_t)-\beta\rho_t
 = (1-\beta)\rho_t + \beta(G_t)$$
So $\beta$ can be thought as a weighing parameter between the previous episode returns ($\rho_t$) and the current return ($G_t$). Now, if the returns have high variance, learning with a higher beta can force the agent to adapt the thresholds very quickly, especially if the thresholds are updated at the end of every episode. This might lead to instability, however, if the updates are done after the convergence of Q-values, then 
this problem will not persist.

So for most practical purposes and given an unknown domain, the easiest solution would be to keep $\beta$ in a range of ~0.001-0.1. For example, for all gym and Atari environments, we used a $\beta=0.1$.

Another option would be to start $\beta$ at a lower value when the rewards have high variance and gradually increase it as the rewards become stable and the agent is more confident about its decisions. This scheduling $\beta$ is another option to use if running a parameter sweep over $\beta$ is very cumbersome. 

\begin{figure*}[h]

\centering
\includegraphics[width=0.48\textwidth]{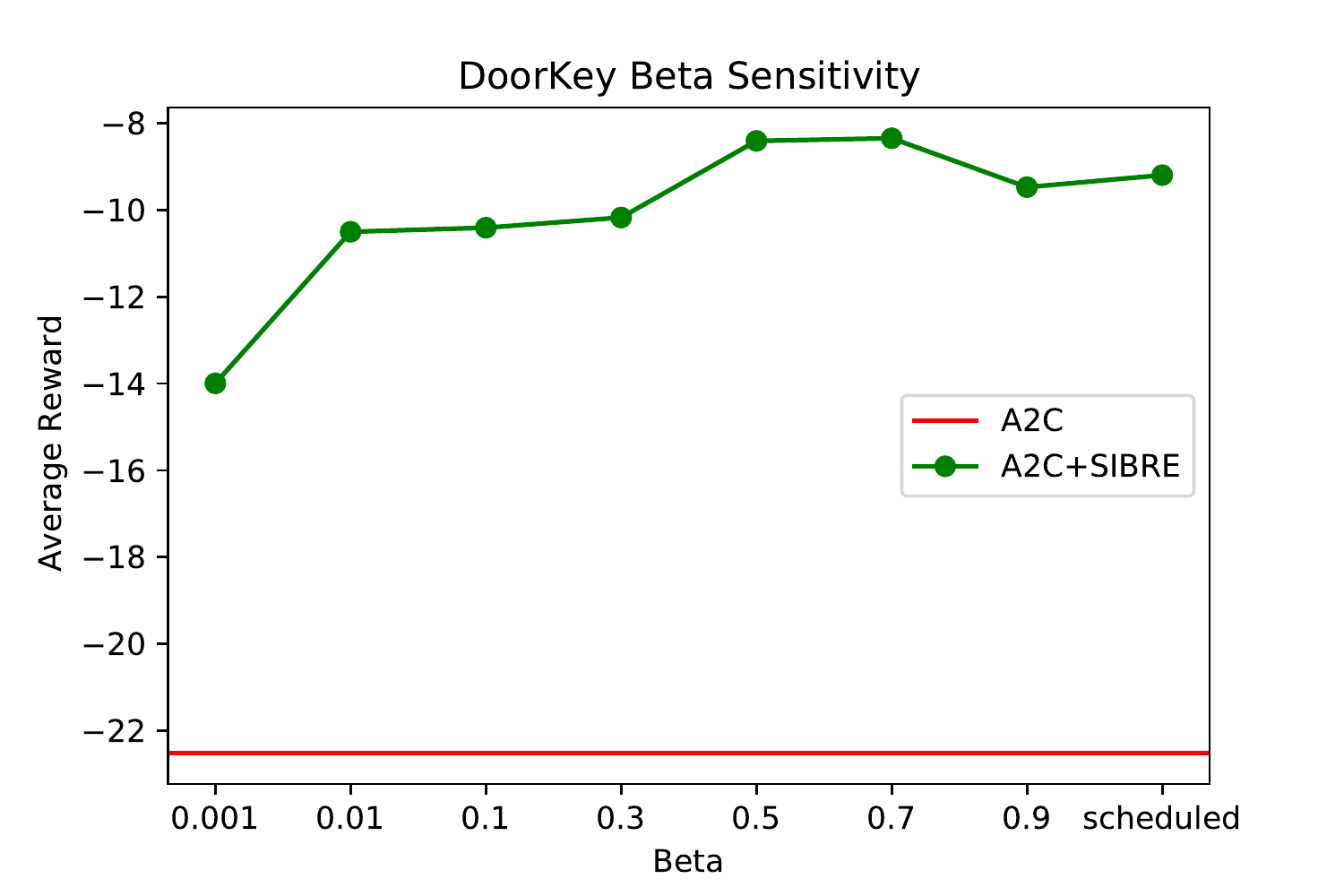} 
\includegraphics[width=0.48\textwidth]{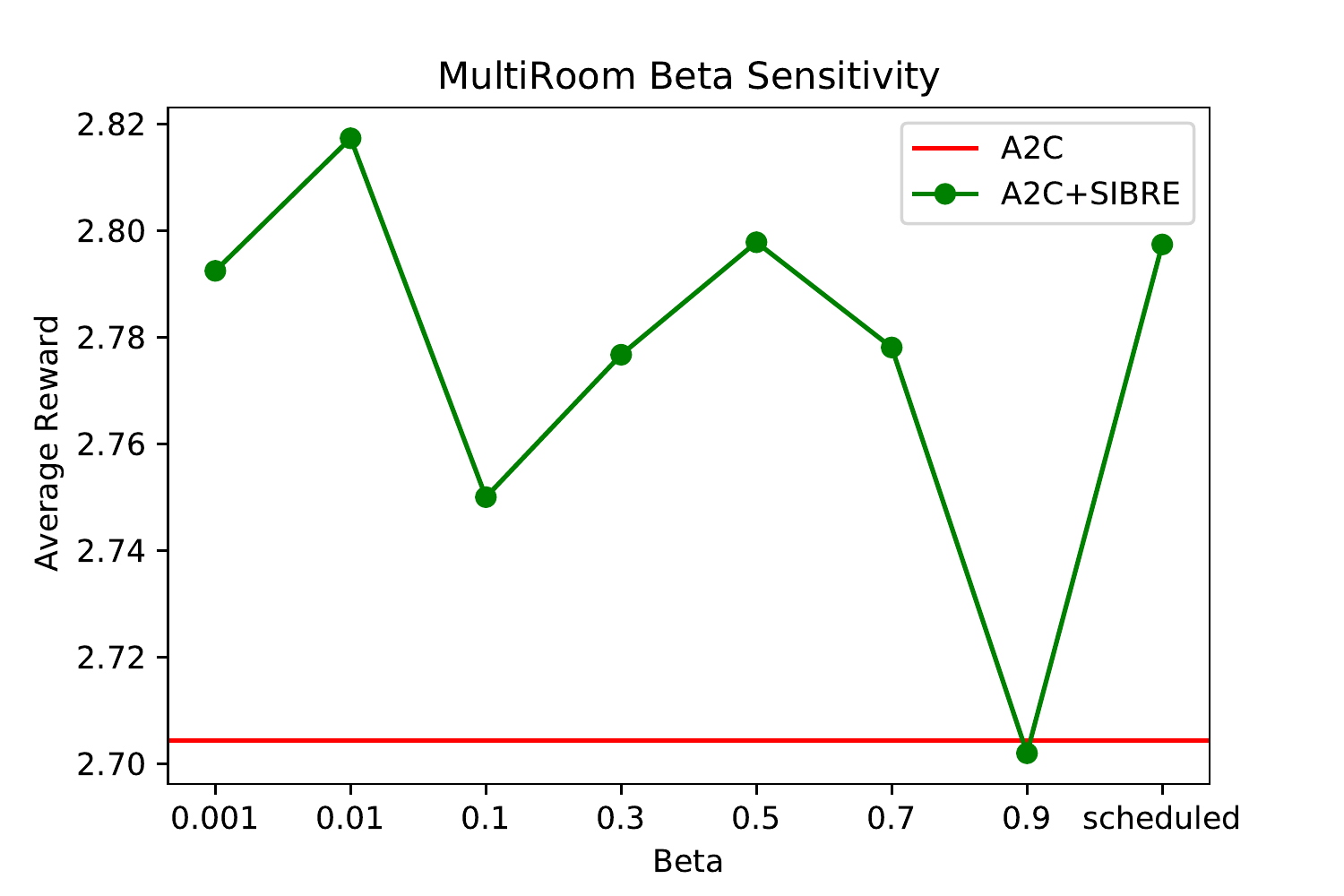} 
\caption{$\beta$ Sensitivity plots for SIBRE on (a) 6x6 Doorkey \& (b) Two Room Environment }
\label{fig:beta}

\end{figure*}

In Figure \ref{fig:beta}, we demonstrate the sensitivity of $\beta$ in DoorKey and Multiroom domains, where we find SIBRE to be fairly robust to varying values of $\beta$ in a proper range  (not too low or too high). Here, too, SIBRE with the suggested range of the values of $\beta$ outperformed the A2C agent without SIBRE.

\subsection{Robustness to Learning Rate}
Aside from the main goal of accelerating learning, we saw that using past performance based adaptive feedback also encouraged learning with smaller learning rates, when the original RL algorithm failed to reach optimal performance. We observed that the gain need not necessarily be only with respect to the performance for a particular learning rate, but also in terms of improving robustness across other learning rates.

\begin{figure*}[h!]
\centering
\includegraphics[width=0.48\textwidth]{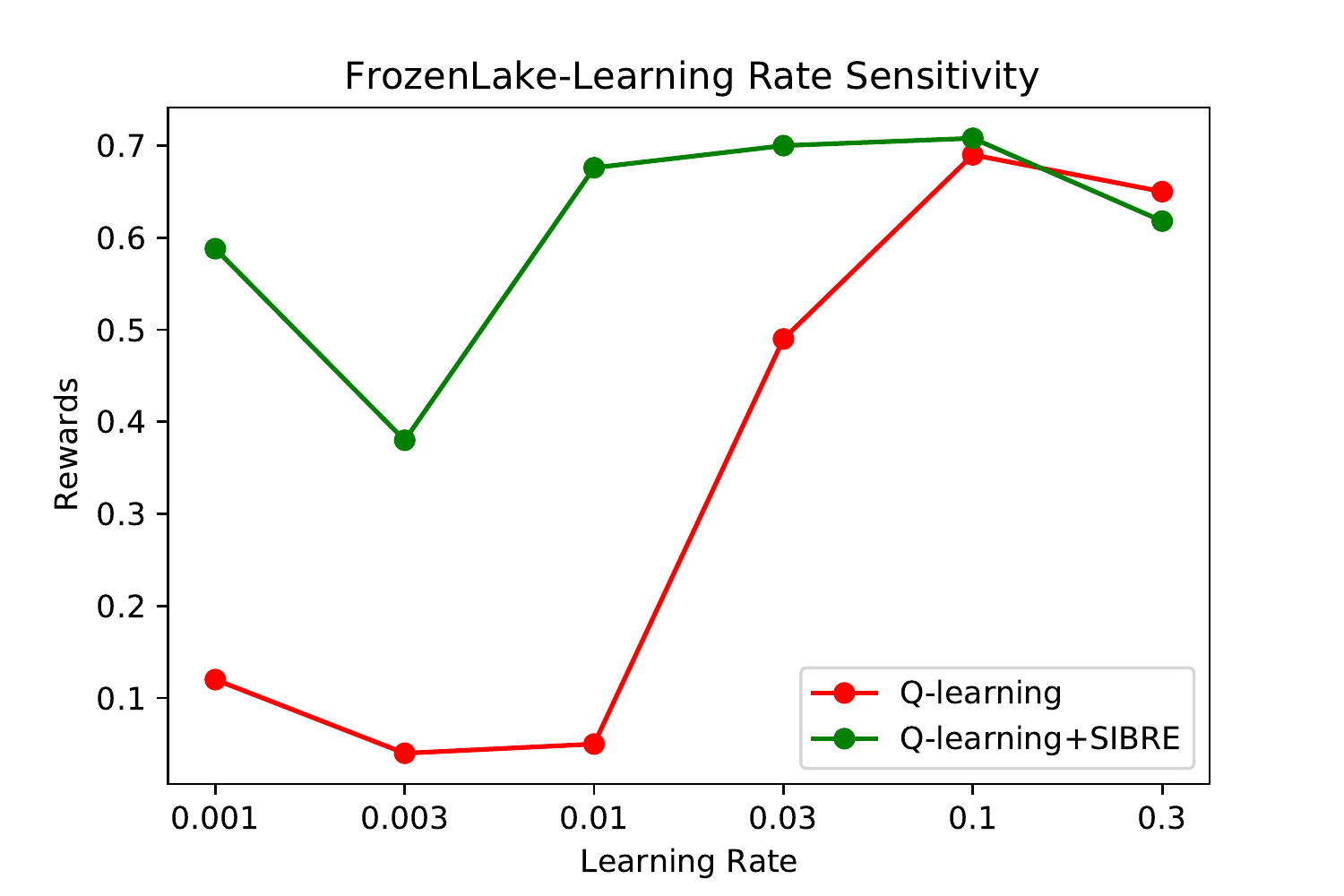} 
\includegraphics[width=0.48\textwidth]{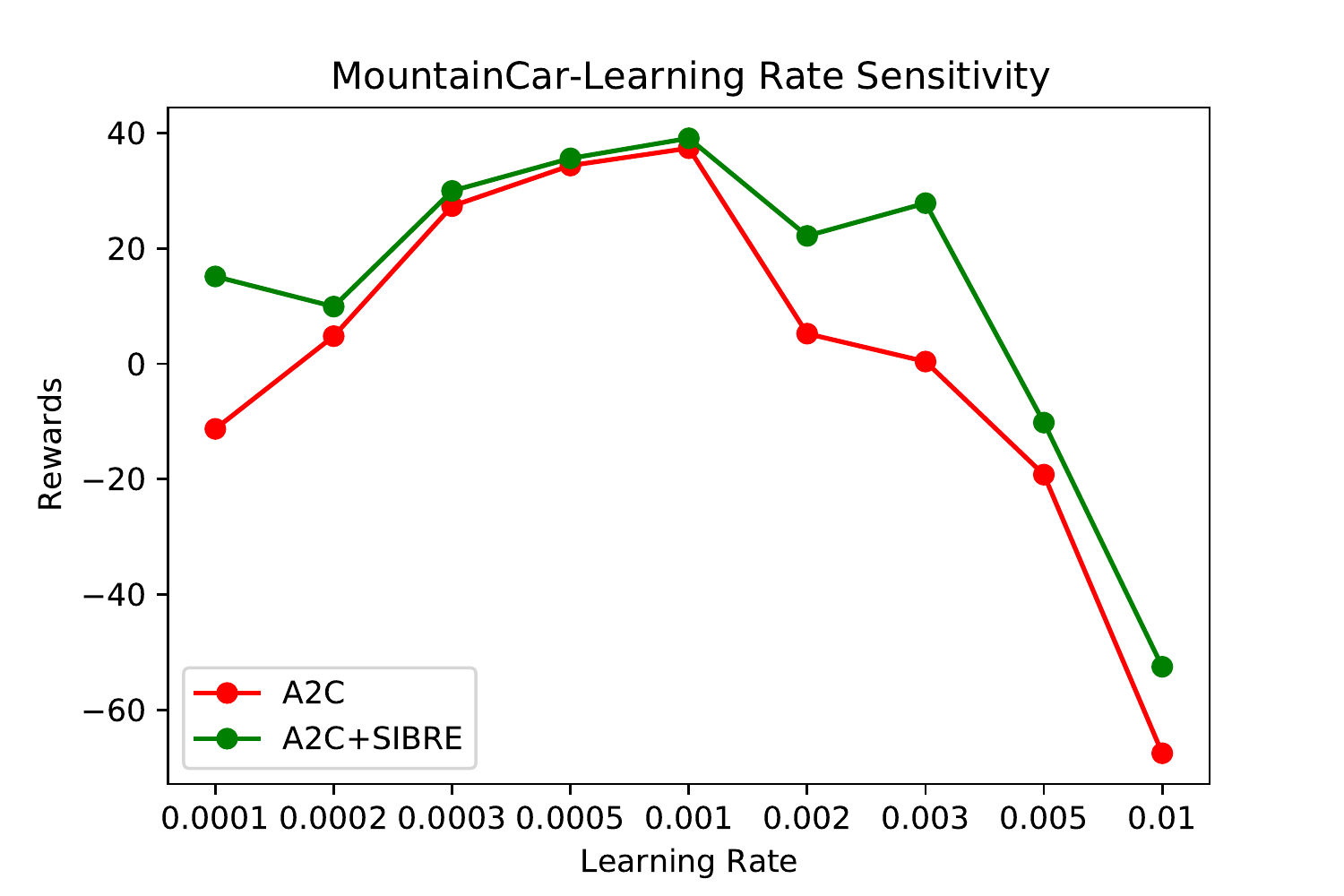} 
\caption{Learning Rate Sensitivity Plots for (a) Q-learning+SIBRE vs Q-learning on FrozenLake (b) A2C+SIBRE vs A2C on MountainCar }
\label{fig:alpha}
\end{figure*}

For FrozenLake, Q-learning performed equally well, with and without SIBRE, however, across various ranges of the learning rate, we found SIBRE to be consistently better. Each of the points in Fig. \ref{fig:alpha} (a) represent the mean reward over 10000 episodes across 10 independent runs. We find Q-learning unable to solve the environment at all for lower learning rates, whereas with SIBRE, it is able to perform much better.

We found similar case in MountainCar as well where A2C+SIBRE was more robust than pure A2C across a variety of learning rates as seen in Fig. \ref{fig:alpha} (b), where each point is average reward across 50 episodes over 10 runs.

\subsection{Faster learning in conjunction with other reward shaping algorithms}\label{sec:sr}
\begin{figure*}
    \includegraphics[width=0.48\textwidth]{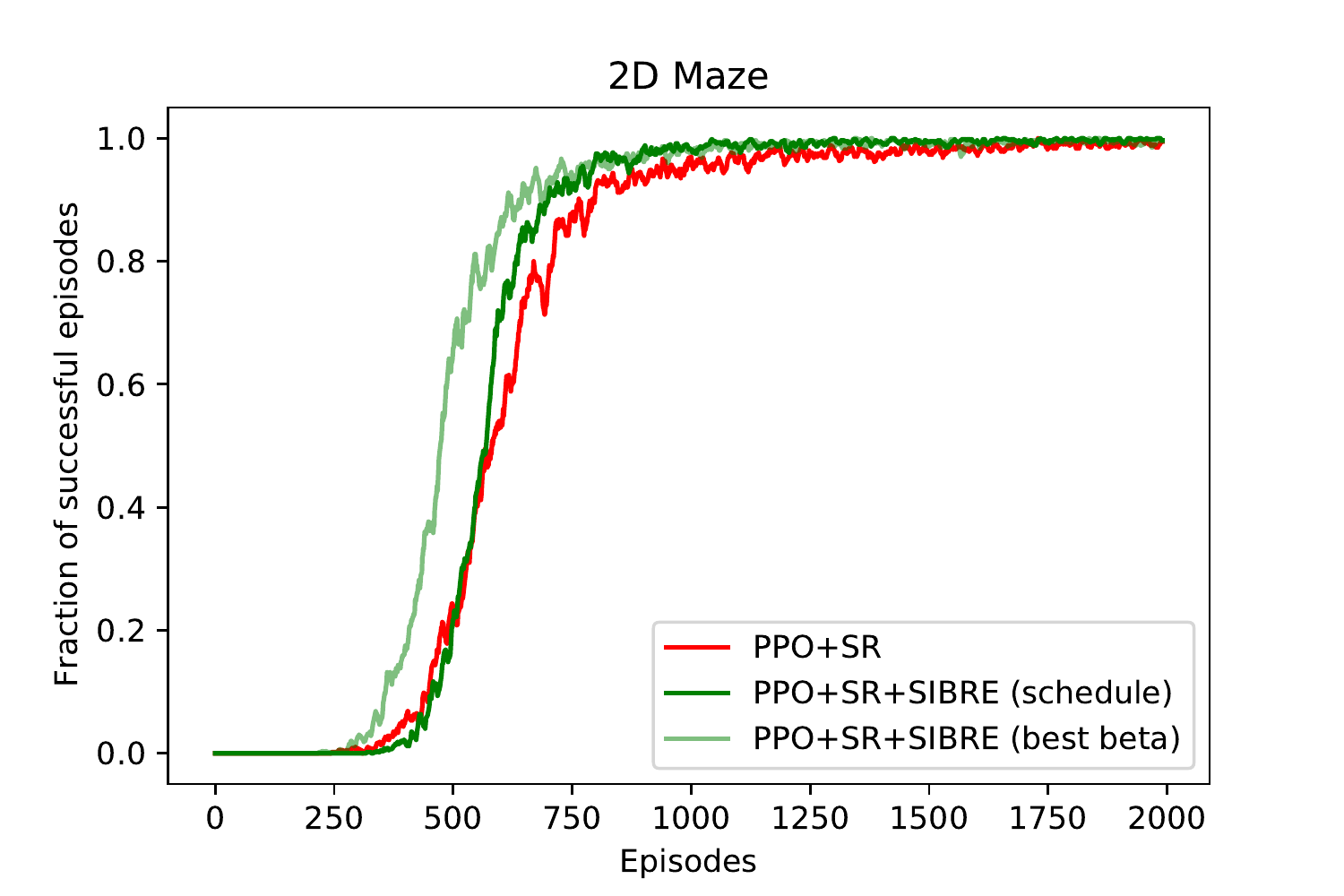}
    \includegraphics[width=0.48\textwidth]{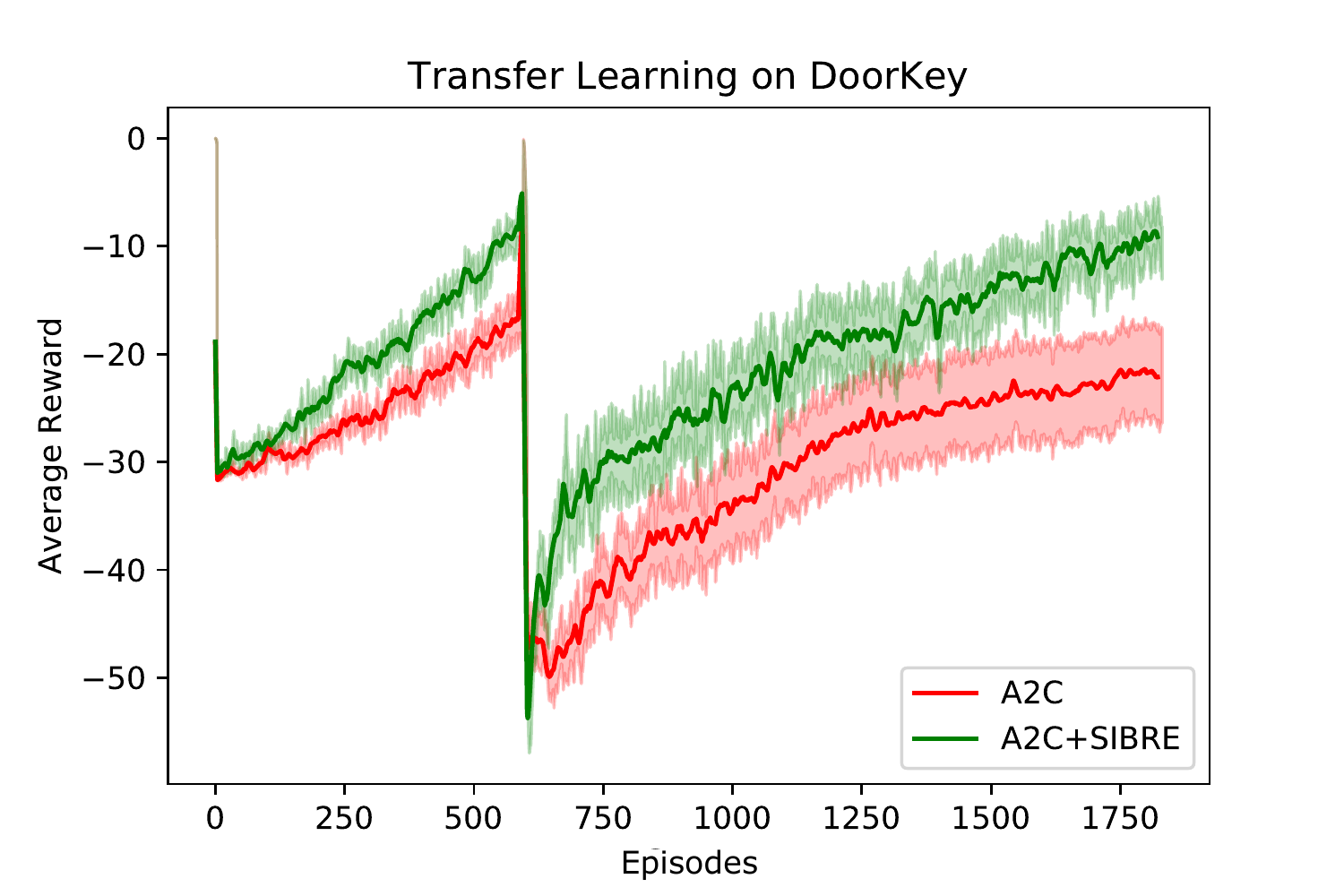}
    \caption{(a) Sibling Rivalry with and without SIBRE on 2D Maze (b) SIBRE+A2C vs A2C on transfer learning from 5x5 to 8x8 grid in DoorKey environment}
    \label{fig:transfer}
\end{figure*}

In this section, we improve upon another reward-shaping algorithm, Sibling Rivalry \citep{NIPS2019_9225}. This algorithm uses a distance based reward metric using two rollouts which encourages exploration particularly in sparse reward domains. However, this can only be used with on-policy algorithms. In our experiment, we used the 2D Maze environment, described in the paper \citep{NIPS2019_9225} and augmented it with SIBRE. 

From our experiments, in this domain too, we found SIBRE to significantly improve performance as seen in Fig. \ref{fig:transfer} (a). We tried with both an optimal beta (obtained from hyper-parameter search) as well as the same beta-schedule used for all the other environments. An important observation is that both the SIBRE agents reach the optimal policy faster than the non-SIBRE agent. This is a significant result considering PPO \citep{schulman2017proximal} fails to solve this problem. This result is particularly encouraging because it demonstrates the flexible usage of SIBRE on top of any modifications made to standard RL algorithms to achieve better performance.

\subsection{Transfer Learning on DoorKey}
SIBRE learns the value of a threshold which it aims to beat after each episode. The threshold encourages it to keep doing better. We believe that once it has learnt the threshold properly, we get optimal performance. When we use the same model to learn on a bigger state-space with same reward structure, the value of the threshold provides a high initial value to beat and this helps in easy transfer of learning. Also, after the transfer, the value of the threshold can also yield information about possible negative transfer across environments.

In DoorKey, we first train on 5x5 grid for 0.8 Million Frames and then transfer to 8x8 grid and train it for further 2.4 Million Frames and the results plotted are averaged over 10 runs. All other hyper-parameters are same as the previous experiment. From Figure \ref{fig:transfer} (b) it is evident that SIBRE enables faster transfer as well as good overall performance. 

\subsection{Extension to Continuing Domain Problems}
\label{sec:continuing}
The basic intuition of SIBRE is the incorporation of information about past episodes into the reward structure so that the agent can learn from it. However, this very definition necessitates SIBRE to be defined only on episodic MDPs.
However, the past reward structures even in continuing problems are not devoid of information and the agent can also use such the past rewards in a better way to improve performance. Instead of updating the rewards after every K episodes, here we update the rewards after every K steps where K is sufficiently large to get a good understanding of the agent performance for the past K steps and hence SIBRE can easily be extended to continuing problems. However, since we are dealing with continuing tasks, we use the discounted return instead of the regular return while calculating the threshold values.

For this experiment, we developed the continuing version of the Cartpole environment, where the agent would receive a reward of 0 at every step and -1 when the pole fell down.

\begin{figure}
\includegraphics[width=0.5\textwidth]{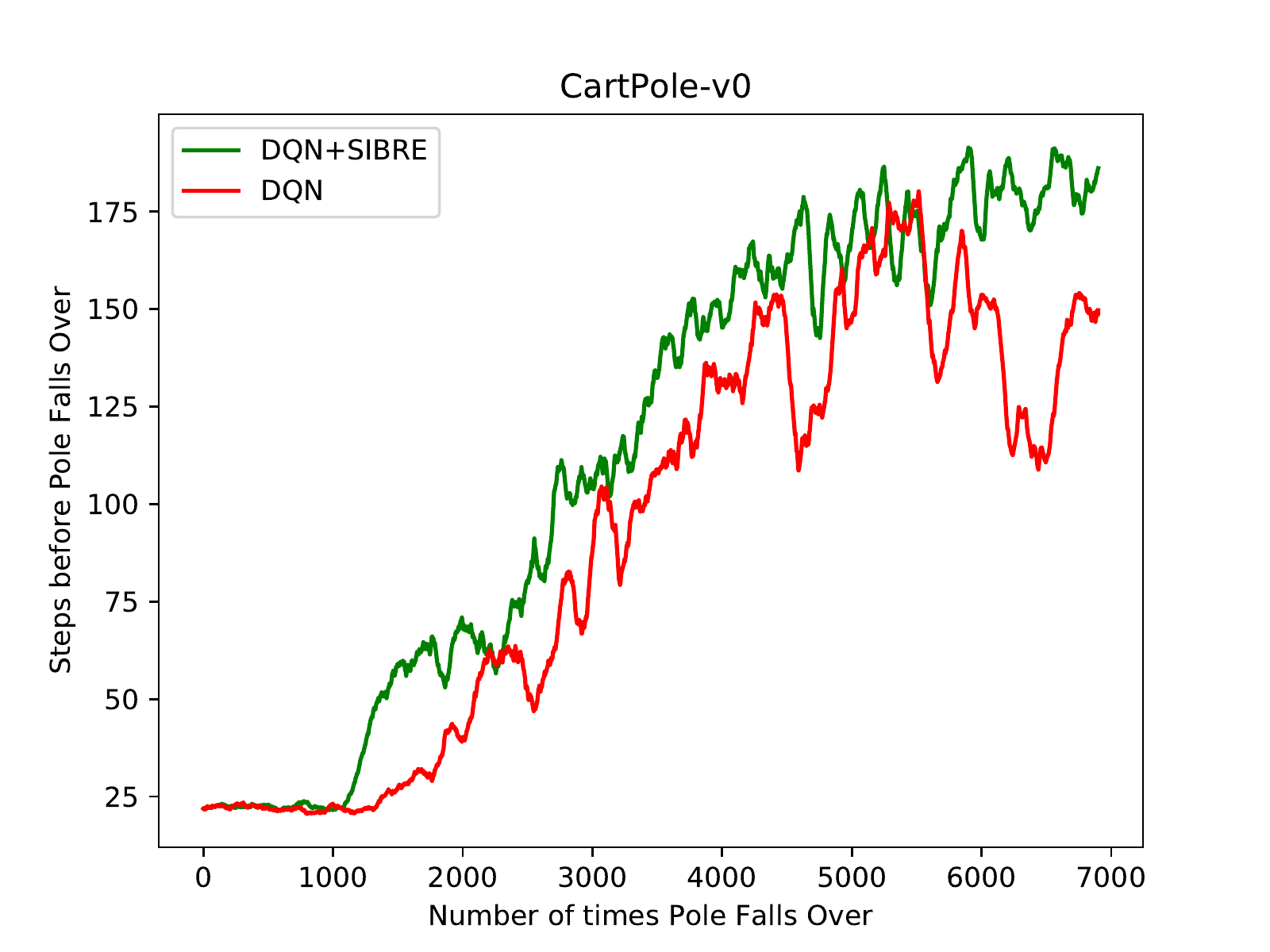} 
\caption{DQN and DQN+SIBRE on continuing Cartpole task}
\label{fig:cartpole}
\end{figure}

Figure \ref{fig:cartpole} portrays the learning curves of DQN (with and without SIBRE) across 10 runs on the continuing CartPole task with SIBRE having a scheduled beta setting with the threshold being updated after every 500 steps. Thus, even in non-episodic MDPs, SIBRE can be effective in accelerating learning.

\subsection{Similar reward shaping in practical problems}

We found three prior studies in literature that used reward shaping in practical applications, that was similar in concept to SIBRE. These studies, however, do not provide a theoretical analysis of the technique. \cite{laterre2018ranked} showed improved results on a bin-packing task using a ranked-reward scheme. A binary reward signal was computed using the agent's own past performance. \cite{khadilkar2018scalable} use a related approach for online computation of railway schedules. They use comparison with a prior performance threshold to define whether the latest schedule has improved on the benchmark. This binary signal (success or failure) is used to train a tabular Q-learning algorithm. \cite{verma2019reinforcement} use improvement with respect to previous objective values as the proxy for total return in a policy gradient framework. While we could not locate any studies that used SIBRE in its precise form for practical problems, the success of similar versions of reward shaping indicates that SIBRE could be useful for such real-world problems where the optimal rewards are not known in advance.



\section{Conclusion and Future Work}

In this work, we showed that an adaptive, self-improvement based modification to the terminal reward (SIBRE) has empirically better performance, both qualitative and quantitative, than the original RL algorithms on a variety of environments. We were able to prove, analytically, that SIBRE converges to the same policy in expectation, as the original algorithms. In particular, we showed on a diverse set of combinations of popular benchmarking environments and original RL algorithms, that the empirical advantages of SIBRE are observed in situations where RL typically has difficulties: when the reward signal is weak, when the returns for different actions are poorly differentiated, when the episodes are long, and when transferring learning from one size of problem instance to another. By using existing historical traces to produce an additional reward signal (improvement over episodes) to aid adaptation, SIBRE is able to learn faster. Also, we showed that SIBRE is able to perform well in continuing, as well as, episodic environments with sparse, as well as, dense rewards and with value-based, as well as, policy-based RL algorithms. 



In future work, we plan to test the potential advantages of using the improvement-based rewards in additional RL environments. We also want to use the technique along with curriculum/transfer learning \cite{curriculum}, especially in partially observed environments, where the state and action space sizes remain the same (local) but the environment complexity (scale, observability, stochasticity, non-linearity) is scaled up. We would also like to evaluate the theoretically provable benefits of SIBRE in terms of \textit{rate} of convergence.


\newpage
\bibliographystyle{ACM-Reference-Format} 
\bibliography{ref}

\newpage
\onecolumn
\section*{Supplementary Material}
\subsection{Additional Experimental Results}
\begin{figure}[h]

\centering
\includegraphics[width=0.48\textwidth]{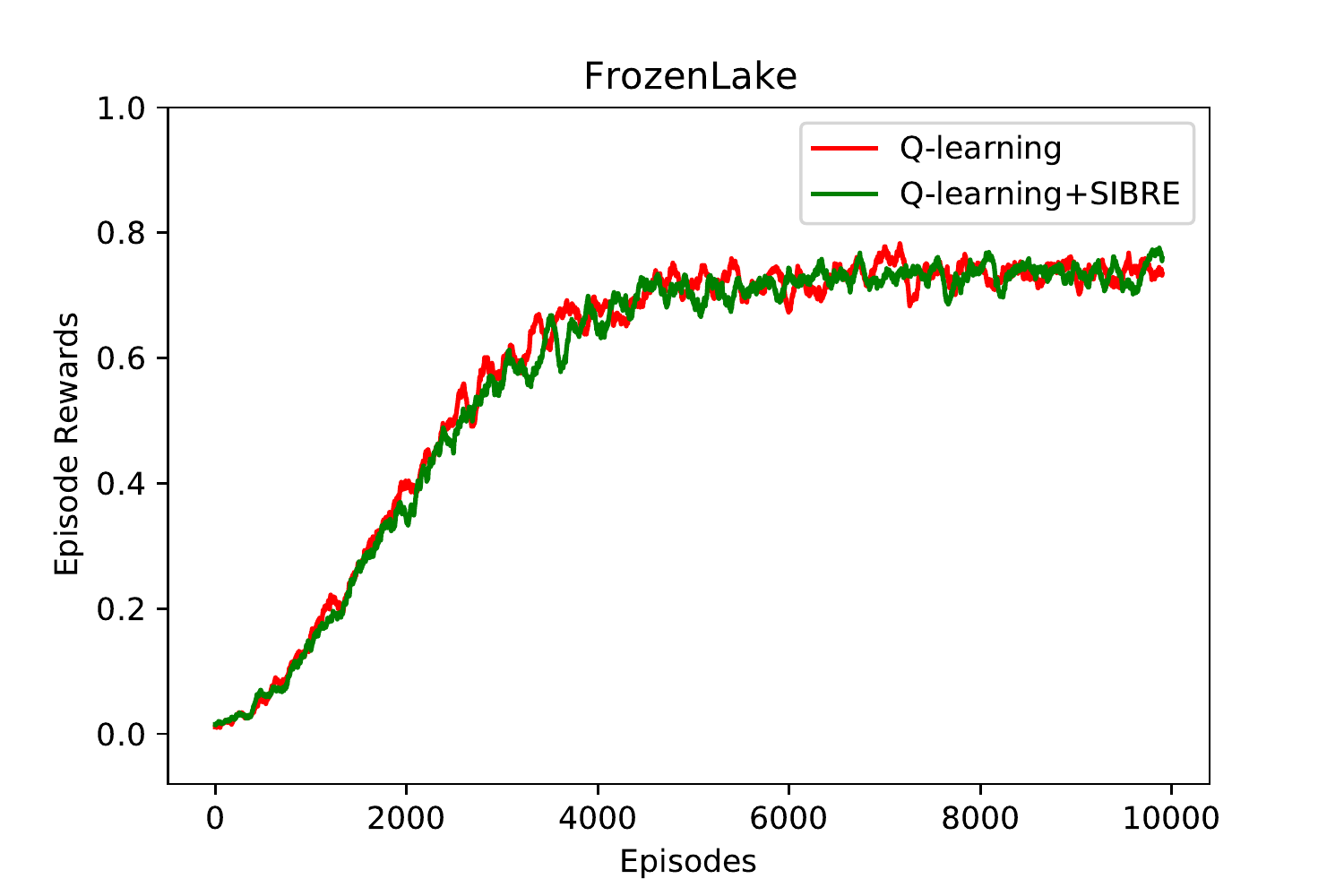} 
\includegraphics[width=0.48\textwidth]{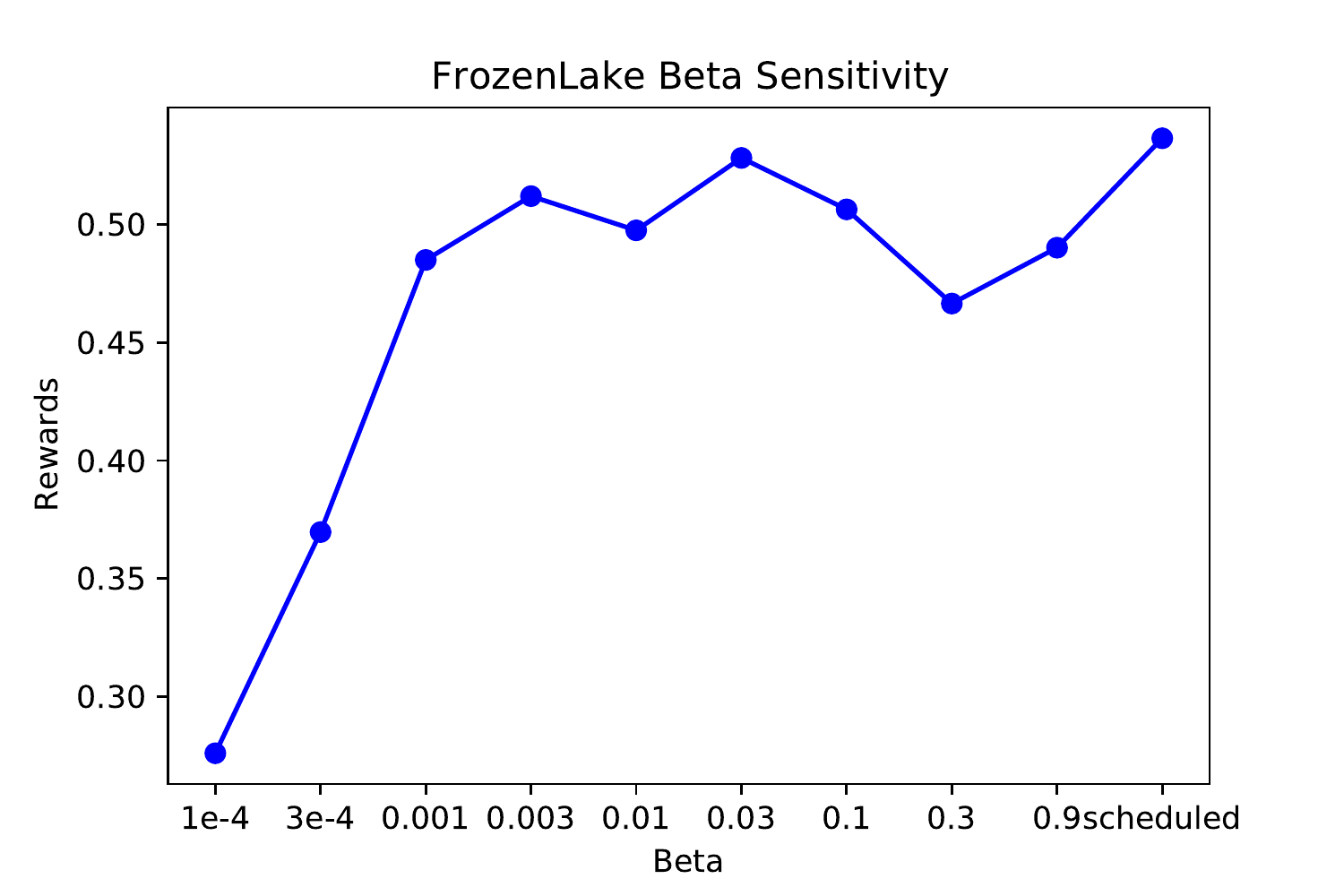} 
\caption{(a) Q-learning with SIBRE vs Q-learning (b) $\beta$ Sensitivity on FrozenLake }
\label{fig:frozenlake}

\end{figure}




Here we plot the additional results like the learning curves on FrozenLake where Q-learning with and without SIBRE performs similarly, as seen in Fig. \ref{fig:frozenlake} (a). In Fig. \ref{fig:frozenlake} (b), the parameter sensitivity of SIBRE with respect to $\beta$ is portrayed. As explained in the main paper, we found the value of $\beta$ to not matter much as long as it was chosen with the proper range. For example, the performance is almost similar across a range of values with scheduled beta doing the best. 

For most other environments, we simply chose $\beta=0.1$ without hyper-parameter tuning, and that worked pretty well.
\onecolumn
\subsection{Hyper-Parameters for all the environments}
\begin{figure*}[h!]
\begin{center}
 \begin{tabular}{|c | c | c | c | c|}
 \hline
 Environment & Algorithm  & Parameters &  Compute  \\  
 \hline \hline
  & &  Frames = 1.8 Million &  \\ 
  & Advantage  & 
 $\alpha = 7\times10^{-4}$  &    \\ 
  Door \& Key & Actor Critic &  entropy coefficient = 0.01 & Google Colab (GPU)  \\ 
  & &   $\gamma = 0.99$ &  \\ 
  & &   $\beta$ = schedule &  \\ 
 \hline
 
 & &  Frames = 1.8 Million &  \\ 
  & Advantage  & 
 $\alpha = 7\times10^{-4}$  &    \\ 
  Multiroom & Actor Critic &  entropy coefficient = 0.01 & Google Colab (GPU)  \\ 
  & &   $\gamma = 0.99$ &  \\ 
  & &   $\beta$ = schedule &  \\ 
 \hline
 
 & &  Turn Limit = 100 &  \\ 
  & &  Episodes = 10000 &  \\ 
  & &  $\epsilon = 1$ &  \\ 
  FrozenLake & Q-learning  & 
 $\alpha = 0.1$  & Google Colab (CPU)  \\ 
  & &  $\epsilon$-decay=0.9/100 steps & \\ 
  & &   $\gamma = 0.99$ &  \\ 
  & &   $\beta$ = schedule &  \\ 
 \hline
 
  & &  Frames = 1 Million &  \\ 
  & &  architecture = 2x32 units  &\\  
  & & $\alpha = 0.001$  & \\   
  CartPole & DQN &  $\epsilon = 1$ & i7-9750H (CPU) \\ 
  & &   $\epsilon$-decay=0.9999 &  \\ 
  & &   $\gamma = 0.99$ &  \\ 
  & &   $\beta$ = schedule &  \\
  
   \hline
 
  & &  Episodes = 50 &  \\ 
  & Advantage &  
 $\alpha = 0.001$  &    \\ 
  MountainCar & Actor Critic &  entropy coefficient = 0.1 & Google Colab (CPU) \\ 
  & &   $\gamma = 0.95$ &  \\ 
  & &   $\beta = 0.1$ &  \\
 


 \hline 
  & &  Frames = 1 Million &  \\ 
  &   & $\alpha = 0.0001$  &    \\ 
  Pong & Rainbow &  buffer size = 100000 & Google Colab (GPU)  \\ 
  & &   $\gamma = 0.99$ &  \\ 
  & &   $\beta = 0.1$ &  \\ 
 \hline

  & &  Frames = 30 Million &  \\ 
    & &   $\beta = 0.1$ &  \\ 
  Freeway & Rainbow & $\gamma = 0.99$  & GTX 1080 FE (Pascal)  \\ 
  & & Parameters same &  \\ 
  & & as Google Dopamine  &    \\
 \hline
 
 & &  Frames = 120 Million &  \\ 
    & &   $\beta = 0.1$ &  \\ 
  Venture & Rainbow & $\gamma = 0.99$  & GTX 1080 FE (Pascal) \\ 
  & & Parameters same &  \\ 
  & & as Google Dopamine  &    \\
 \hline

  & &  Episodes = 25000 &  \\ 
  & &  Max Length = 50 &  \\ 
  2D Maze & Proximal Policy &  $\beta = 0.002$ & 20-core CPU \\
  & Optimization & $\alpha = 0.001$  &  \\
  & &   $\gamma = 0.99$ &  \\ 
 \hline

\end{tabular}
\end{center}
\end{figure*}

\end{document}